\documentclass[preprint,12pt]{elsarticle}

\usepackage{color}
\usepackage{amsmath}
\usepackage{amssymb}
\usepackage{array}
\usepackage{url}
\usepackage{verbatim}
\usepackage{graphicx}

\usepackage{enumerate}
\usepackage{enumitem}
\usepackage{float}
\usepackage{bbm}
\usepackage{booktabs}
\usepackage{caption}
\usepackage[normalem]{ulem}
\usepackage{algorithm}
\usepackage{algpseudocode}

\newcommand{\NEW}[1]{\textcolor{black}{#1}}

\begin{document}

\begin{frontmatter}

\title{Combining Transformers \\with Natural Language Explanations}

\author[inst1]{Federico Ruggeri\corref{cor1}}
\ead{federico.ruggeri6@unibo.it}
\author[inst2]{Marco Lippi}
\author[inst1]{Paolo Torroni}

\cortext[cor1]{Corresponding Author.}

\affiliation[inst1]{organization={Department of Computer Science and Engineering},
addessline={viale Risorgimento 2},
city={Bologna},
country={Italy}}

\affiliation[inst2]{organization={Department of Sciences and Methods for Engineering},
addessline={via Amendola 2},
city={Reggio Emilia},
country={Italy}}

\begin{abstract}
Many NLP applications require models to be interpretable. However, many successful neural architectures, including transformers, still lack effective interpretation methods. A possible solution could rely on building explanations from domain knowledge, which is often available as plain, natural language text. We thus propose an extension to transformer models that makes use of external memories to store natural language explanations and use them to explain classification outputs. 
We conduct an experimental evaluation on two domains, legal text analysis and argument mining, to show that our approach can produce relevant explanations while retaining or even improving classification performance.
\end{abstract}

    
\end{frontmatter}

\section{Introduction} \label{sec:introduction}

In recent years, deep learning models have produced stunning results in a wide variety of domains. In natural language processing (NLP), transformer-based architectures like BERT~\citep{devlin2018bert} and more recently large language models like GPTX, Chinchilla, LLaMa, PaLM and BLOOM~\citep{foundationmodels}
set a new standard in many tasks%
~\citep{chernyavskiy2021transformers}.
In spite of the welcome leap in performance, however,
a typical criticism transformer architectures share with most deep learning models is their lack of interpretability. Sure, the attention mechanism~\citep{AttentionSurvey} could offer cues as to how to interpret the behavior of such models. Nevertheless, whether attention could be meaningfully used as an analysis tool, especially from the perspective of a layman or end-user,  is 
a matter of discussion~\citep{Jain2019,Wiegreffe2019}.

\begin{figure*}[!t]
    \centering
    \includegraphics[width=0.8\linewidth]{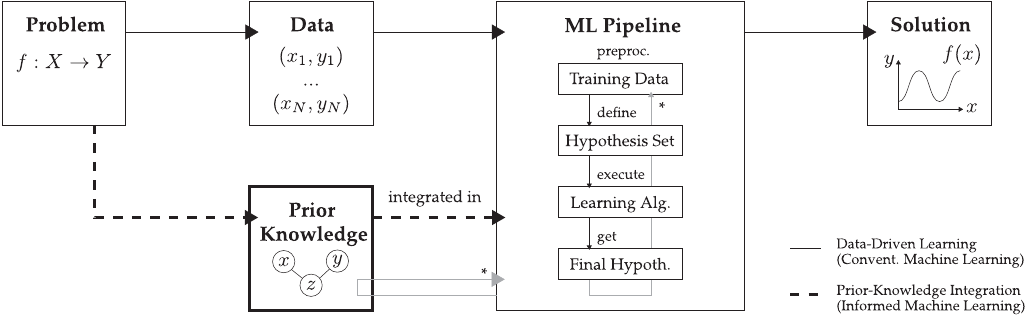}
    \caption{Information flow in informed machine learning. From~\citep{von_Rueden_2021}}
    \label{fig:knowledge_integration}
\end{figure*}

A way to make deep networks more interpretable could be to inject them with recognizable, readable knowledge elements, as illustrated in Figure~\ref{fig:knowledge_integration}~\citep{von_Rueden_2021}. 
%
A few interesting proposals in this direction rely on data augmentation, model architectural biases, regularization constraints, and retrofitting. 
For instance, objective constraints could be used to view background knowledge as a set of differentiable and possibly learnable rules \citep{hu2020harnessing,hu-etal-2016-deep}.
However, these are structured methods that require translating expert knowledge into some knowledge representation formalism. Typically, that implies a costly abstraction process and a compromise in terms of conceptual simplifications imposed by the limitations of the representation language.
To illustrate the point, let us consider the case of unfair clause detection in online terms of contracts~\citep{lippi2019claudette}. Take the following clause:

\begin{quote}
    \textit{We reserve the right to modify or terminate the service or your access to the service for any reason, without notice, at any time, and without liability to you.}
\end{quote}

A legal expert may consider such a clause potentially unfair, because of the powers it gives the provider:  to \textit{unilaterally remove consumer content}, and to \textit{unilaterally terminate} the services. 
In a legal analytics scenario~\citep{ashley2017artificial} where the identification of unfair clauses is done automatically, a system's output of ``potential unfairness'' could be explained by the distribution of attention mass on specific segments of text. However, this in itself is not the type of explanation that a legal expert would provide to a non-expert: it shows a characteristic of the inference mechanism but does not offer any explanatory abstractions. 
If asked to explain why these issues would make a clause potentially unfair, a legal expert could instead resort to general principles  (\textit{rationales}) such as \textit{unilateral change} and \textit{unilateral termination}, which define aspects of unfairness in this particular domain of law. These may be explained, even to a layman or end-user, as follows:

\begin{description}
\item[Unilateral Change:] \textit{the provider has the right for unilateral change of the contract, services, goods, features for any reason at its full discretion, at any time.}
\item[Unilateral Termination:] \textit{the contract or access may be terminated for any reason, without cause or leaves room for other reasons which are not specified.}
\end{description}

Here, the identification of relevant rationales is the result of an abstraction process carried out by the legal expert in an effort to explain an opinion. Similarly, an automatic system for unfair clause detection could use a shortlist of legal rationales to justify a prediction of unfairness, as they would define the relevant principles leading to such a prediction.


%

Endowing the classifier with the ability to use legal rationales could be seen as a form of knowledge injection. However, if such a knowledge injection required translating the legal principles into a formal language of sort, other issues would arise.
For starters, a structured conceptual mapping of the reasoning behind such justifications would be extremely time-consuming and hampered by referral actions, common sense motivations, and other well-known issues in NLP~\citep{ruggeri2021detecting}. 
Moreover, it would require defining applicability requirements, conditions that must hold to ensure the correct usage of the knowledge provided. Conversely, an architecture able to exploit knowledge expressed in plain English would circumvent many of such problems. 
Moreover, textual information may be readily available or easier to obtain than structured knowledge.

Following a recent trend in NLP, one could think of using large language models and designing suitable prompts to accompany a query with relevant knowledge. 
\NEW{However, recent studies~\cite{shen2023,zhou2023} have shown that LLMs do not necessarily represent the best solution for tasks that involve memory injection. On the contrary, it has been argued that encoder-only models are still state-of-the-art in information extraction task, where decoder-based LLMs still lag behind.} In fact, it would be difficult to verify that the knowledge injected by prompting is actually used, notwithstanding prompt scalability issues with complex legal questions and long lists of possible rationales, let alone the risk of hallucinating. Besides, \NEW{the experimental design suffers from being highly-dependent from the prompts used for the study, and } the post-processing required to make sure that the generated answer satisfies certain quality requirements would suffer from the same limitations of structured conceptualization discussed earlier on.

\NEW{
In~\cite{von_Rueden_2021}, the authors discuss the possible role of knowledge integration for improving interpretability, something not yet sufficiently explored in current literature. In~\citep{ruggeri2021detecting}, we move a step in the direction of enhancing interpretability by knowledge integration, by showing that legal rationales expressed in the form of unstructured plain text can be used to explain classifications of unfairness in contract clauses. 
}
%
%
Here we extend our initial proposal to accommodate transformers, and to improve the applicability and scalability of this approach. In this way we can successfully accommodate hundreds of explanations on regular computing hardware, making it possible to tackle non-trivial domains. Our approach could be described as \textit{learning from textual descriptions}, and it resembles certain human learning processes\NEW{~\cite{kintsch1986,vandenbroek2010}: comprehending the text and incorporating the comprehended information in the reader's background knowledge.} We validate our method on two domains: legal text classification and argument mining. Importantly, we achieve this result without hampering the classification performance.


%
 Our main contributions are:
\begin{enumerate}
  \item an approach to extend transformer architectures with an external memory containing natural language sentences that can be used as explanations;
  \item  an efficient, publicly available implementation,\footnote{Public code and data repository:~\url{https://github.com/lt-nlp-lab-unibo/bert-natural-explanations/}} which thanks to smart sampling strategies is able to scale up to memories with hundreds of elements on regular computing hardware; 
  \item a thorough validation conducted on two challenging tasks: unfairness detection in consumer contracts and claim detection in argument mining.
  \end{enumerate}

The paper is structured as follows. Section~\ref{sec:problem_formulation} presents our memory-en\-hance\-ment approach, introduces
strong supervision, and proposes different attention-based sampling strategies to address scalability.  Section~\ref{sec:experimental_setup}  describes the experimental setting, whereas Section~\ref{sec:results} presents and discusses results and in-depth memory interaction analysis. Section~\ref{sec:related_work} discusses related work. Section~\ref{sec:conclusions} concludes.

\section{Related Work} \label{sec:related_work}
Memory networks and extensions of transformer architectures have been hot topics in recent NLP research.
A recent trend regarding combining memory-augmented architectures with transformer architectures views the memory component as auxiliary support for efficient learning. More precisely, a specific effort is dedicated to solving the well-known problems of handling long-term dependencies, as well as processing long sequences. One common idea is to reduce the search space by progressively storing intermediate input representations in the memory component. Notable examples are Transformer-XL~\citep{DBLP:journals/corr/abs-1901-02860}, Reformer~\citep{DBLP:journals/corr/abs-2001-04451}, LongFormer \citep{DBLP:journals/corr/abs-2004-05150} and MemFormer~\citep{DBLP:journals/corr/abs-2010-06891}. In these architectures, the memory component is directly inserted as an intermediate layer in the transformer architecture, at the multi-head attention level, in order to condition token selection.

In a similar fashion, but with different goals, transformers have been augmented with an external memory block to efficiently deal with data. This enables handling complex input sequences like videos~\citep{DBLP:journals/corr/abs-1912-08226} by overcoming the natural sequential information flow and learning global-level dataset information. In the latter scenario, global and local attention mechanisms have been combined by means of learnable parameter vectors~\citep{DBLP:journals/corr/abs-1907-01470,DBLP:journals/corr/abs-2006-11527} that account for word meanings in different parts of the dataset~\citep{DBLP:journals/corr/abs-2008-01466}.

In contrast, we do not propose an architectural modification of transformer architecture. Instead, we leverage the general-purpose architecture of memory-augmented neural networks and use transformers as a possible implementation of one of the constituting blocks of such an architecture. Furthermore, our role of memory is different. Rather than exploiting a support memory for storing past information, we make use of the memory to introduce domain knowledge with the aim of making a model interpretable.

Another frequent use of the memory block is for external knowledge integration. Transformers have been directly applied as advanced input encoding methods in traditional memory-augmented architectures for information retrieval in dialogue systems~\citep{DBLP:journals/corr/abs-1811-01241,DBLP:journals/corr/abs-2010-05740,DBLP:journals/corr/abs-2107-07566,DBLP:journals/corr/abs-2107-07567},  question-answering~\citep{DBLP:journals/corr/abs-1909-09696} and aspect-based sentiment analysis~\citep{8918438}. In these scenarios, the memory block is employed to store texts that are specific to each individual input example~\citep{DBLP:journals/corr/RajpurkarZLL16,DBLP:journals/corr/abs-1811-01241}. Alternatively, external memories have been used to encode knowledge bases in the form of high-level metadata like knowledge graphs~\citep{gaur2021semantics}, commonsense knowledge~\citep{verga2020facts}, and semantic links between entities~\citep{9044058,DBLP:journals/corr/abs-1911-00172}. In the latter proposals, each memory slot encodes a single word or entity tuple with the goal of expanding the contextual information of words and thus improving the learning process. 

In our approach, knowledge is unstructured, plain text without any structured semantic information. Moreover, knowledge has a different role from the external knowledge used in question-answering and reading comprehension tasks, where it contains information required to solve the task like, for instance, the answer to a query. There, memory sampling strategies are not applicable unless some arrangements take place like allowing unanswerable questions if the text containing the answer is not provided.

Indeed, with knowledge in the form of sentences in natural language, we cover several scenarios of different nature, spanning from additional contextual information to general task-oriented constraints. For instance, the memory could contain textual descriptions of the task itself and, thus, we could formulate the task of learning from class descriptions, intuitively resembling human learning~{\citep{DBLP:conf/emnlp/WellerLGP20}}. 

Among the many approaches to explaining black-box models~{\citep{explainability}}, a recent trend in AI is using natural language explanations. For example, Lertvittayakumjorn and Toni define explanations as text fragments in the input text that are most relevant to a prediction~{\citep{lertvittayakumjorn2019human,lertvittayakumjorn2021explanation}}.
\NEW{\citep{hartmann-sonntag-2022-survey} recently summarize several approaches that fall withing the paradigm of \textit{learning from explanations}.
In particular, these approaches rely on extractive explanations (i.e., highlights), free-text explanations, and semi-structured explanations that lie in between extractive and free-text explanation types.
These explanations are provided in curated human-annotated corpora that may support multiple explanation types, such as e-SNLI~\cite{camburu2018}, WinoWhy~\cite{zhang2020}, CoS-E~\cite{rajani2019explain}, and ERASER~\cite{deyoung2020}.
According to the taxonomy of \citep{hartmann-sonntag-2022-survey}, we use free-text explanations stored in external memories given a priori as a pool of possible justifications.
In particular, the explanations are not generated as an output of the model but rather retrieved and integrated during the learning process and may describe a whole class of examples rather than being sample-specific.
Lastly, in our work, we do consider knowledge representations of different types like legal rationales and class-related examples, both integrated with the same role of explaining a model's predictions.}


Alternative attempts at building interpretable NLP models have been mainly focusing on the attention mechanism~{\citep{AttentionSurvey}}.
Although attention can be used to interpret the behavior of deep neural networks~\citep{explainability}, whether it can also be considered a proper \emph{explanation} tool is still an open debate, which hinges on the definition of explanation itself.
In fact, while attention provides a possible interpretation of a network's behavior, it is not possible to guarantee whether that interpretation is the right one~\citep{Jain2019,Wiegreffe2019}.
Thus, we can consider attention weights to be an explanation if we define them as a \emph{plausible}, but not necessarily \emph{faithful}, reconstruction of the decision-making process~\citep{Rudin,Riedl}.
It can be argued that attention is not \emph{consistent} with other explanation methods~\citep{Jain2019}, but the same characteristic has been observed in many other popular explanation tools~\citep{ordercourt}.
Attention has also been used as a means to integrate knowledge in models by specifically training it to select relevant features~\citep{AAAI1816485,attention-rationales} or to model an auxiliary task~\citep{linguistically-informed-attention,dual-attention}.
On a slightly different perspective, the ``interpretability illusion'' of BERT-related models has been described as the phenomenon for which individual neurons in BERT do not always show a human-interpretable meaning~\citep{bolukbasi2021interpretability}.

\section{Approach} \label{sec:problem_formulation}

The design choices behind our memory-augmented transformers architecture are motivated by our intended use of external textual knowledge.
Past research on deep learning models tackled the problem of handling external content by introducing external memory blocks the model could interact with in a differentiable way~\citep{SukhbaatarSWF15}. Such \emph{memory-augmented neural models} enabled complex reasoning tasks like reading comprehension and question answering and opened a new path in the context of meta-learning~\citep{yin2020metalearning}.
Therefore, they seem an ideal candidate architecture for the integration of natural language explanations. 
However, the main purpose of our extension of transformers is not to improve classification performance, but to generate explanations in the form of grounding to elements of a textual knowledge. Accordingly, the knowledge stored in our memory is not required, strictly speaking, to correctly address the task (as, for example, in question answering). It is instead an auxiliary collection of pieces of information that can be consulted to perform reasoning and to make the neural model interpretable. 

\subsection{Memory-Augmented Transformers} \label{sec:memory_augmented_transformers}

\begin{figure*}[!bt]
    \centering
    \includegraphics[width=0.80\linewidth]{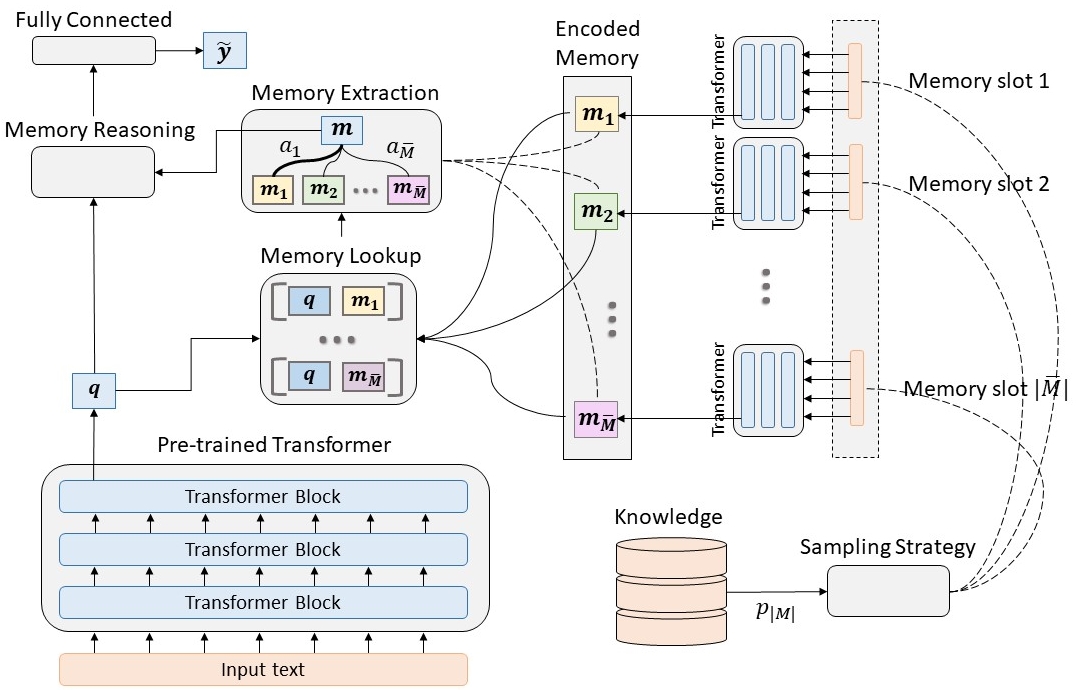}
    \caption{Memory-augmented transformer architecture. $M$ denotes the whole memory, while $\bar{M}$ is a memory subset after a possible sampling step. ${\vert} \cdot {\vert}$ denotes cardinality.}
    \label{fig:membert}
\end{figure*}

The main building block of our architecture is a transformer-based model. In our experiments, we considered BERT~\citep{devlin2018bert} and DistilBERT~\citep{sanh2019distilbert}, a distilled version of BERT that achieves competitive performance while limiting the overall computational burden. However, our approach is general and is not restricted to these two architectures.
Memory-augmented neural network architectures \citep{weston2014memory,graves2014neural} (MANNs) extend neural networks with an external memory block that supports reasoning. We denote our memory-augmented transformer models as \emph{memBERT} and \emph{memDistilBERT}. The memory component is loaded with relevant background textual knowledge. In particular, the memory $M$ comprises several slots, i.e., \textit{memory slots}, each storing a single element of the loaded textual knowledge.
Our architecture (see Figure~\ref{fig:membert}) follows the typical structure of MANNs. Given an input text and a memory component with $\vert M \vert$ memory slots:

\begin{enumerate}[label=\roman*)] 
\item  a transformer layer encodes input text and memory slots, producing a set of sentence embeddings; 
\item a memory-lookup layer compares memory slots with an input text, producing a set of similarity scores; 
\item a memory extraction layer converts such similarity scores into attention scores $\{a_1, \dots, a_{M}\}$ and uses them to compute a single memory summary embedding vector via a weighted sum of the memory slots; 
\item a memory reasoning layer uses the memory summary embedding vector to update the input embedding vector. For instance, a simple form of input update consists in summing input and memory summary embedding vectors together; 
\item a fully-connected layer uses the updated vector as an input for the classification task. 
\end{enumerate}

Being $N$ the number of training examples and $C$ the set of classes, all models are trained to minimize standard cross-entropy loss:

\begin{equation} \label{eq:clf_loss}
    \mathcal{L}_{CE} = - \sum_{i=1}^{N} \sum_{c=1}^{C} p(y_i = c) \log p_\theta (y_i = c)
\end{equation}
\noindent where $\theta$ are the model parameters.

In this setting, there is no information about which is the correct memory slot to be applied to a given example. In the literature, this is known as Weak Supervision (WS).

\subsection{Guiding Memory Interaction with Strong Supervision}

The concept of WS is important since not always memory slots can be explicitly associated with individual training examples (e.g., this might be costly with regard to the time required from an expert, but also intrinsically challenging). This is a very general setting. Yet, if such annotation information is provided, we can regularize the model in the training phase by focusing on specific target memory slots. 
This procedure is formally known as Strong Supervision (SS) and it has been widely explored in memory networks since their introduction~\citep{SukhbaatarSWF15}. SS can assume different formulations.
In our experimental scenarios, it is sufficient to enforce a preference for just one of the (possibly multiple) memory slots associated with a given input example. Following~\citep{ruggeri2021detecting} 
we exploit SS in the form of a max-margin loss. 
We define target memory slots as those that have been linked to the input example during the annotation phase. We introduce a penalty term that enforces target memory slots to have a higher  similarity score with respect to the input $x$ than to the remaining slots, up to a $\gamma$ margin:

\begin{equation}
\begin{aligned}
    \mathcal{L}_{SS} ={} & \frac{1}{N} \sum_{n=1}^{N} \frac{1}{Z} \sum_{m_+, m_-} \mathcal{L} (m_+, m_-) \\
    \text{s.t.} & \quad Z ={} \vert M^n_+ \vert \vert M^n_- \vert \\
    \text{and} & \quad m_+ \in M^n_+ \, m_- \in M^n_-
\end{aligned}%
\end{equation}

\begin{equation}
\begin{aligned}
    \mathcal{L} (m_+, m_-) & ={} \max \Big (0, \gamma - \tilde{s}(x, m_+, m_-) \Big ) \\
    \text{s.t.} \quad & \tilde{s}(x, m_+, m_-) ={} \sigma (s(x, m_+)) \\
    & + \sigma (s(x, m_-))
\end{aligned}%
\end{equation}


%
\noindent where for the $n$-th example $M^n_+$ is the set of target memory slots for a given input example $x$, $M^n_-$ is the set of non-target memory slots for the $n$-th input example, $(m_+, m_-)$ is a target/non-target memory slot pair for $n$, $\sigma$ is the activation function and $s(\cdot, \cdot)$ is the similarity function. 
The hyper-parameter $\gamma$ controls the intensity of the SS constraints, while trading-off classification performance and model interpretability: large values of $\gamma$ would basically turn the preference ranking induced by the hinge loss into a classification loss.
Overall, the loss function for memBERT and memDistilBERT with SS is:
\begin{equation}
    \mathcal{L} = \mathcal{L}_{CE} + \mathcal{L}_{SS}
\end{equation}

\subsection{Efficiently Handling Large Memories with Sampling}

In the architecture described so far, efficiently handling large memories is particularly challenging for two main reasons. First, as memory size grows, i.e., the number of memory slots increases, the number of irrelevant (if not noisy) memory slots for a given input example inevitably grows as well. Second, the complexity of the memory-related layers (steps (ii)--(iv) in Section~\ref{sec:memory_augmented_transformers}) also increases with the memory size.

To this end, approximate solutions or sampling-based methods have been explored,
trading-off performance with efficiency~\citep{chandar2016hierarchical}. Typical solutions mainly exploit simple comparison strategies, like dot product similarity, that are easily executed in parallel and approximated.

In our case, we aim to define a more complex memory lookup phase. Therefore, we allow neural-based similarities and focus on 
an efficient sampling strategy to achieve scalability.
Memory sampling brings about both benefits and, possibly, issues. Reducing the overall memory size enables large knowledge integration when dealing with complex deep neural models like transformers. Moreover, sampling eases the memory integration process from the point of view of noise compensation, i.e., filtering out irrelevant memory slots. In a similar fashion to curriculum learning~\citep{curriculum}, sampling reduces complexity by considering a smaller memory, which, in turn, induces a limitation of spurious memory lookup operations.
This is also reflected at training time, where, in each batch, the model sees a different sampled memory $\bar{M}$, thus avoiding always focusing on specific slots, which could possibly head to overfitting.
Nonetheless, sampling inherently introduces variance, therefore compatibility issues with other strategies like SS might arise as well. To compensate for both problems, we define smart sampling strategies that prioritize memory slots according to the observed input examples. To this end, similarly to prioritized experience replay (PER)~\citep{SchaulQAS15}, we introduce priority-based sampling strategies that progressively take into account the importance of the added information with respect to both input examples and task objectives. In particular, to efficiently deal with large memory sizes, we operate at the batch level by sampling a shared memory $\bar{M}$. Then, depending on the given strategy, priority weights $p_{m_i}$ associated with each memory slot $m_i$ are updated and available for the next batch step.
More precisely, we adopt the same priority definition of PER, where the temporal difference error is replaced by a custom importance assignment function:
\begin{equation} \label{eq:per}
    p_{m_i} = (w_{m_i} + \epsilon)^\alpha
\end{equation}
\noindent where $w_{m_i}$ is the importance weight of the $i$-th memory slot $m_i$, $\epsilon$ is a sufficiently small real value that avoids zero-based exponentiation and $\alpha$ controls the priority degree. Intuitively, $\alpha$ equals 0 produces all priorities equal to 1, thus defining a uniform sampling strategy.
Lastly, sampling is applied both at training and at inference time, when learned priority weights are employed to sample the memory.
\NEW{In particular, memory importance weights $w_{m_i}$ are only learned during training where information about ground-truth labels is provided.
At inference time, the memory sampling distribution $p_{|M|}$ is fixed.}
Algorithm~\ref{algo:training_sampling} and Algorithm~\ref{algo:inference_sampling} describe the general training and inference procedures with memory sampling, respectively.
The introduced strategy variants differ in how priority weights are updated at each batch step.

\begin{algorithm}[!t]
\caption{Training Procedure with Memory Sampling}
\label{algo:training_sampling}
\begin{algorithmic}[1]
    \Require Data $D = \{ (\textbf{x}_n, \textbf{y}_n) \}_{n=1}^{N}$, \newline Memory weights distribution $p_{\vert M \vert}$, \newline Model weights $\theta$
    \State Initialize model weights $\theta$ = $\theta^0$
    \State Initialize memory weights $p_{\vert M \vert}$ = $p^0_{\vert M \vert}$
    \While{Stopping criteria}
    \State Get minibatch $B = \{(\textbf{x}_b, \textbf{y}_b)\}_{b=1}^{{\vert} B {\vert}} \subset \mathcal{D}$
    \State Sample memory $\bar{M}$ from $M$ using $p^{k-1}_{{\vert} M {\vert}}$
    \State Compute $\mathcal{L}(x {\vert} \bar{M})$ on $B$
    \State Update $\theta^k$ using any optimizer
    \State Compute $w^k_{{\vert} M {\vert}}$ (e.g., Eq. \ref{eq:strat:attention}-\ref{eq:strat:lg}) 
    \State Update $p^k_{{\vert} M {\vert}} = (w_{m_i} + \epsilon)^\alpha$ (Eq.~\ref{eq:per})
    \State Normalize $p^k_{{\vert} M {\vert}}$
    \EndWhile
    \Ensure Trained weights distribution $p_{{\vert} M {\vert}}$
\end{algorithmic}
\end{algorithm}

\begin{algorithm}[!t]
\caption{Inference Procedure with Memory Sampling}
\label{algo:inference_sampling}
\begin{algorithmic}[1]
    \Require Data {$D = \{ (\textbf{x}_n) \}_{n=1}^{N}$}, \newline Trained weights distribution {$p_{{\vert} M {\vert}}$}
    \While{$B \ne \varnothing$}
    \State Get minibatch, {$B = \{(\textbf{x}_b)\}_{b=1}^{{\vert} B {\vert}} \subset \mathcal{D}$}
    \State Sample memory {$\bar{M}$  from $M$ using $p_{{\vert} M {\vert}}$}
    \State Get model predictions {$\{(\tilde{\textbf{y}}_b)\}_{b=1}^{{\vert} B {\vert}}$}
    \EndWhile
    \Ensure Predictions and sampled memory {$\{ (\tilde{\textbf{y}}_n, \bar{\textbf{M}}_n) \}_{n=1}^{N}$}
\end{algorithmic}
\end{algorithm}

\subsubsection{Uniform Sampling}

As a baseline strategy, we consider uniform memory sampling. At each batch, each memory slot has the same probability of being selected. Priorities are kept fixed during training. Formally, being $\vert M \vert$ the memory size, we define the probability of sampling the $i$-th memory slot $m_i$ as $p_{m_i} = \frac{1}{\vert M \vert}$, with $i = 1, \dots, \vert M \vert$.

\subsubsection{Priority Sampling}

Intuitively, uniform sampling does not consider the importance of each memory slot. 
In order to do that, at training time, each individual input example should modify the underlying sampling distribution by giving more priority to those memory slots that have been reputed as useful. We explore two different formulations of priority that ground the notion of usefulness to a specific architectural property of the model.
More precisely, we consider the memory attention layer as the major indicator of the importance of each memory slot with respect to the given input example, considering two variants: (i) the first one relies exclusively on the attention value of each memory slot, whereas (ii) the second formulation also takes into account the contribution of each memory slot to the primary training objective $\mathcal{L}_{CE}$ (Eq.~{\ref{eq:clf_loss}}).

\paragraph{\textbf{Attention-Based Priority Sampling}} \label{sec:attonly_sampling}

In this formulation, the priority associated with each memory slot during sampling is based only on the corresponding attention value computed by the memory lookup layer, given an input example. Specifically, given a memory slot $m_i$, the attention score computed by the memory lookup layer for a given input text is the importance weight $w_{m_i}$ (see {Eq.~\ref{eq:per}}). At the batch level, we inherently require to aggregate each memory slot importance weight into a single scalar value. Formally, given a batch $\mathcal{B}$ of examples and a memory slot $m_i$, we need to aggregate the importance weights $w^b_{m_i}$ for each example $b \in \mathcal{B}$ to compute the importance weight $w_{m_i}$ and, consequently, the priority weight $p_{m_i}$. Additionally, since the textual knowledge encoded in memory slots only targets a subset of examples (i.e., one specific class of examples in a classification setting), we require to aggregate the importance weights $w^b_{m_i}$ only for such a subset of examples in a given batch. To do so, we compute a binary mask specific to that subset of examples and perform a masked aggregation of attention values for each memory slot accordingly. We denote this method as \textit{masked average aggregation}.
This strategy accounts for noise introduced by input examples for which no target memory slot exists. For instance, in the case of a binary classification task where memory contains valuable information for the positive class only, we should limit sampling priority updates to positive examples to avoid spurious memory selections.

Formally, the attention-based priority sampling defines each memory slot's importance as its summary attention value in the batch.
\begin{equation} \label{eq:strat:attention}
    w_{m_i} = \frac{\sum^{\vert B \vert}_j a_{j, i}}{\sum^{\vert B \vert}_j \mathbbm{1}_{y_j \in \textbf{Y}_+}}
\end{equation}
where $a_i$ is the attention value attributed to memory slot $i$, $\vert B \vert$ is the batch size, and $\textbf{Y}_+$ is the set of positive examples within the batch.

Note that this strategy is purely similarity-based and does not consider each memory slot's contribution to the task-specific objective. In other terms, it operates under the assumption that each knowledge descriptive concept contained in the memory should have high semantic and syntactic similarity with its associated input examples. Therefore, we argue that, in cases where input examples have high similarity with unrelated memory slots, the proposed approach might not be appropriate, as it would aim to perform the task at hand by combining the input with irrelevant information.

\paragraph{\textbf{Loss Gain Priority Sampling}}

In order to render priority sampling aware of the usefulness of each memory slot to the task itself, we consider an additional variant that directly takes into account the classification loss. In particular, we consider the loss difference between the model architecture and the one without the memory layer. By doing so, we are able to estimate the informative gain introduced by the external knowledge with respect to the task. The larger the difference, the larger the impact of the memory. To associate a specific priority value to each memory slot $m_i$, we weigh the computed difference by the corresponding attention value $m_i$. We consider the same reduction operation strategy adopted for attention-based priority sampling.
Formally, this loss gain priority sampling can be formulated as follows:
\begin{equation} \label{eq:strat:lg}
    w_{m_i} = \frac{ \sum_j^{{\vert} B {\vert}} a_{j, i} \exp^{(\mathcal{L}_{CE}(x_j) - \mathcal{L}_{CE}(x_j {\vert} \bar{M}))}}{\sum^{{\vert} B {\vert}}_j \mathbbm{1}_{y_j \in \textbf{Y}_+}}
\end{equation}
\noindent where $a_i$ is the attention value attributed to memory slot $i$, $\bar{M}$ is the sampled memory, $\mathcal{L}_{CE}(x)$ and $\mathcal{L}_{CE}(x {\vert} \bar{M})$ are the cross-entropy loss obtained without any interaction with the memory, or with the introduction of the sampled memory $\bar{M}$, respectively. The exponential guarantees a positive priority while awarding positive differences with high values and down-weighting negative differences.

\section{Experimental Setup} \label{sec:experimental_setup}

To test our approach, we consider two distinct classification scenarios. In the first, we extend the work in~\citep{ruggeri2021detecting} focusing on the legal domain for unfair clause detection in online Terms of Service, by employing explanations given by legal experts as domain knowledge. In the second scenario, we explore the argument mining case study of claim detection by considering evidence as the source of knowledge, following the assumption that class correlation might ease the classification task.

These scenarios motivated some architectural choices regarding the usage of memory. Typically, when memory slots contain connected segments of text, steps $(ii)-(iv)$ described in Section~\ref{sec:problem_formulation} are repeated multiple times to capture sequential information. However, since our memory slots contain independent fragments, we run these steps just once. In the MANN jargon, this means setting the number of hops to 1. 

Both our case studies have input examples associated with multiple target memory slots and examples associated with none. For example, a negative result (a clause that is not unfair) has no associated justification. 
For these reasons, we adopt sigmoid-based rather than softmax-based attention scores.

For each case study, we carried out hyper-parameter calibration following standard procedures (see Section~I in the Supplementary Material).

Concerning memory sampling, we consider sampling a small memory $\bar{M}$ with memory size ${\vert} \bar{M} {\vert} \ll {\vert} M {\vert}$ for efficiency reasons. A small memory can reduce potential noise derived from unrelated memory slots and maintain low demand for computational resources. We evaluated different memory sizes in a preliminary experimental setting to observe their effects on model performance. We eventually arbitrarily picked a small ${\vert} \bar{M} {\vert}$ for both tasks based on how frequently a memory slot is associated with an input example.
\NEW{Nonetheless, we report a sensitivity analysis of $\lvert \bar{M} \lvert$ in Section 3 of the Supplementary Material.}

\subsection{Knowledge as Class Descriptions: Unfairness Detection}

Online Terms of Service (ToS) often contain potentially unfair clauses to the consumer. 
Unfair clause detection in online ToS is a binary classification task where each clause is labeled as either fair (negative class) or unfair (positive class)~\citep{lippi2019claudette}. Multiple unfairness categories result in multiple binary classification tasks. Recent proposals~\citep{ruggeri2021detecting} address this problem by using legal expertise in the form of \textit{rationales}. Because rationales define the unfairness category, they can be used as explanations for the unfairness predictions.
In our setting, the input text is a clause to be classified, and the shared memory contains all the possible legal rationales concerning reasons for unfairness for positive examples.

\subsubsection{Data}
The dataset originally published in~\citep{ruggeri2021detecting}, named ToS-100, contains 100 documents. It defines legal rationales for five unfairness categories following a multi-label perspective (i.e., a clause can be unfair according to more than a single category). These categories are: \begin{description} \item[\textbf{A}] arbitration on disputed arising from the contract;
\item[\textbf{CH}] the provider's right to unilaterally modify the contract and/or the service; \item[\textbf{CR}] the provider's right to unilaterally remove consumer content from the service, including in-app purchases;\item[\textbf{LTD}]  liability exclusions and limitations; \item[\textbf{TER}] the provider's right to unilaterally terminate the contract. \end{description} To assess the ability of our architecture to cope with limited training data, we randomly extracted from ToS-100 30 documents, while ensuring that each document contained at least one example per unfairness category. We name this dataset ToS-30.
Table~\ref{tab:tos_dataset_statistics} reports some statistics on ToS-30, showing the number and percentage of unfair clauses per category and the amount of corresponding legal rationales defined by experts. More technical details concerning the definition of both the unfairness categories and the rationales can be found in~\citep{ruggeri2021detecting}.

\begin{table}[tb]
\centering
\small
\caption{Dataset statistics for ToS-30. For each category of clause unfairness, we report the number and percentage of unfair clauses and the number of rationales. \label{tab:tos_dataset_statistics}}
\resizebox{\columnwidth}{!}{%
\begin{tabular}{ccccc}
    \toprule
    Clause type & No. Unfair clauses & \% Unfair clauses & No. Rationales \\
    \midrule
    Arbitration  (A)                        & 45 & 0.8 & 8 \\
    Unilateral change (CH) 			        & 89 & 1.7 & 7 \\
    Content removal (CR)	                & 58 & 1.1 & 17 \\
    Limitation of liability (LTD)  			& 161 & 3.0 & 18 \\
    Unilateral termination (TER)    	    & 121 & 2.3 & 28 \\
    \bottomrule
  \end{tabular}%
  }
\end{table}

\subsubsection{Setting}

As in~\citep{ruggeri2021detecting}, we consider each unfairness category individually. For a given unfairness category, we consider as positive examples those that have been labeled as being unfair for the given category, while the remaining sentences are viewed as not unfair.\footnote{The dataset labeling distinguishes between potentially unfair and clearly unfair clauses, but following~\citep{ruggeri2021detecting} we merged these categories together.}
We consider the following neural baselines: (i) 2-layer CNN network; (ii) 1-layer bi-LSTM network; (iii) the MANN model introduced in~\citep{ruggeri2021detecting}, which computes the sentence embedding as an average over word embeddings. Neural baselines use 512-dimensional embeddings, except for MANN which, following~\citep{ruggeri2021detecting}, uses 256-dimensional embeddings. The embeddings for these three baselines are learnt from scratch.\footnote{We also investigated the use of pre-trained word embeddings like GloVe~\citep{DBLP:conf/emnlp/PenningtonSM14}, but we did not observe a performance improvement.}
For transformer-based architectures, we fine-tune pre-trained embeddings.
For memory lookup, we use a single-layer MLP with 32 units for MANN and transformer-based implementations. We consider a sigmoid activation function for the MLP layer to account for examples belonging to the negative class for which the memory should not be used~{\citep{ruggeri2021detecting}}. The reasoning layer is a concatenation of input and memory summary embedding vectors. Overall, the trainable parameters of the memory-augmented transformer architecture are: (i) the pre-trained transformer weights; (ii) the MLP-based memory lookup; (iii) the fully connected layer of the final classification block. For (i), we remark that we share the parameters using the same transformer architecture when encoding textual inputs and memory slots.
Additionally, models are further regularized with $L^2$ penalty with $10^{-5}$ weight, dropout rate in the $[0.5, 0.7]$ range \NEW{for baseline models\footnote{Section 2 in the Supplementary Material reports examples of training optimization stability.} and $0.2$ for transformer-based models}, and early-stopping based on the validation F1-score, with patience equal to 10 for \NEW{transformer-based} models\footnote{We initially trained for 3-4 epochs as a standard practice, but we immediately saw that more epochs were required to achieve significant performance.} and to 50 for other neural baselines. We use Adam, with $10^{-3}$ learning rate for optimization \NEW{for baselines and $5e^{-6}$ for transformer-based models}. When considering SS, we experiment with margin $\gamma = 0.5$.
Regarding memory sampling, we experiment with a sampled memory size $\vert \bar{M} \vert$ of 5. This size is around 80-20\% of the total memory size $\vert M \vert$, depending on the unfairness category (see Table~\ref{tab:tos_dataset_statistics}). This is reasonable as an input clause may be associated with multiple legal rationales.
For evaluation, we employ a 10-fold cross-validation procedure with multi-start with 3 repetitions, choosing the best one according to the validation F1-score.

\subsection{Knowledge as Supporting Facts: Claim Detection} \label{sec:setup:claim-detection}

The second classification scenario is an argument mining task: claim detection~\citep{lippi2016argumentation}. \NEW{In this context, arguments are claim/evidence pairs. Claims are short fragments of text expressing a possibly controversial position about a given topic. Evidence is typically made by longer sentences or paragraphs that report on results of studies, known facts, expert opinions, etc. Evidence is used to support claims. We frame claim detection as a binary classification task where each input sentence can be either identified as containing a claim or not.  We treat evidence as background knowledge. 
We use the evidence-to-claim link annotation as strong supervision to guide the memory selection component.} 

\subsubsection{Data}
We consider a portion of the IBM2015 argumentative dataset built in the context of the Debater project~\citep{rinott2015show,slonim2021autonomous}. The dataset consists of a collection of Wikipedia pages, grouped into topics. \NEW{The annotation procedure carried out by IBM  assumes that claims and evidence are annotated with respect to a given topic.}
In our study, we selected the four topics with the largest amount of claims and evidence. Subsequently, we defined the set of evidence as the model memory. \NEW{We consider an explanation to be correct if the selected memory slots correspond to evidence linked to the given claim in the original ground truth data set}. Due to how argumentative texts were annotated, there are cases in which a single sentence may contain both a claim and evidence or, more seldom, evidence spans through multiple sentences and incorporates a claim. Indeed, these cases hinder the quality of selected data, but due to the low amount of such spurious sentences, we do not consider further pre-processing steps and use the dataset as is. Table~\ref{tab:ibm2015_statistics} reports details about the selected subsets of topics, focusing on their argumentative content. We remark that evidence in the IBM2015 dataset are typically statements extracted from studies or facts established by experts: it is thus a reasonable assumption to have a list of such items available to support claim detection.
\NEW{The following reports an example of a topic, claim and its linked evidence found in the IBM2015 corpus.}

\begin{description}
\item[Topic] \textit{This house would embrace multiculturalism.}
\item[Claim] \textit{Indigenous peoples have the right to the dignity and diversity of their cultures, traditions, histories and aspirations which shall be appropriately reflected in education and public information.}
\item[Evidence] \textit{In October 2007, former Australian Prime Minister John Howard pledged to hold a referendum on changing the constitution to recognise indigenous Australians if re-elected. He said that the distinctiveness of people's identity and their rights to preserve their heritage should be acknowledged.}
\end{description}

\begin{table}[tb]
    \centering
    \small
    \caption{Topics extracted from the IBM2015 dataset, with the associated number of evidence, \NEW{claims, and total number of samples}. For claims, we also report the percentage with respect to the overall dataset size, which corresponds to the frequency of the positive class.\label{tab:ibm2015_statistics}}
    \begin{tabular}{ccccc}
    \toprule
    Topics & No. Evidence & No. Claim & Claim Ratio & No. Samples \\
    \midrule
    1 & 130 & 113 & 4.2\% & 2,699 \\
    2 & 239 & 201 & 4.0\% & 4,988 \\
    3 & 578 & 288 & 3.8\% & 7,642 \\
    4 & 642 & 374 & 3.5\% & 10,546 \\
    \bottomrule
    \end{tabular}
\end{table}

\subsubsection{Setting}
We frame claim detection as a sentence-level binary classification task,  where each sentence can either be identified as containing a claim or not. To  evaluate the benefits of added knowledge, we consider incremental subsets of topics, spanning from 1 up to 4. However, differently from the legal domain case study, the memory content dramatically increases as the number of topics gets larger, from 130 with a single topic up to 642 with four topics. Additionally, the number of claims associated with each evidence is very low (range 1-5). Thus, the correct usage of memory and strong supervision is extremely challenging. Moreover, the increased memory size prohibits using all the given knowledge with transformer-based architectures due to hardware limitations. Hence, the need to adopt a sampling strategy.
Specifically, we experiment with a sampled memory size $\bar{M}$ of 10. This size is around 2-8\% of the total memory size, depending on the number of considered topics (see Table~\ref{tab:ibm2015_statistics}).
Given the low amount of samples, models are evaluated via a $k$-fold cross-validation procedure (we chose $k$=4).
All hyper-parameters and training configurations are identical to those described for the ToS-30 setting, except for the MANN baseline, which now uses 50-dimensional embeddings.

\subsection{Memory Analysis}

Evaluation of memory usage cannot be reduced to task-specific metrics only, like classification's F1. Certainly, one may hope for increased performance. However, our main motivation behind knowledge integration is interpretability, hence our evaluation should mainly take that into account. To that end, we base our evaluation on suitable memory usage statistics, under the assumption that memories contain human-interpretable, natural language justifications. Human studies could be an alternative, more direct measure of interpretability, which we may consider in future developments.
 
As an initial step, we consider several memory-related metrics already introduced in~\citep{ruggeri2021detecting}. These metrics evaluate memory usage from several perspectives, covering typical information retrieval analysis concerning top-K memory ranking and purely memory-oriented statistics like target coverage. Such metrics will also be employed to analyze the behavior of sampling in domains where the full memory size is prohibitively large.
For space issues, we only select a subset of such metrics which we consider most informative:
\begin{itemize} 
\item \textbf{Memory Usage (U)}: percentage of examples for which memory is used.
\item \textbf{Coverage (C)}: percentage of positive examples for which (at least one) memory slot selection is correct.
\item \textbf{Coverage Precision (CP)}: percentage of examples using memory, for which (at least one) memory slot selection is correct.
\item \textbf{Precision at K (P@K)}: percentage of positive examples for which at least one correct target memory is retrieved in the first K positions.
\item \textbf{Mean Reciprocal Rank (MRR)}: average of the reciprocal memory ranks, by considering the highest-ranking correctly retrieved target memory.
\end{itemize}

One problem is how to define an appropriate activation threshold $\delta$, i.e., the minimum value to consider a memory slot has been used by the model. While in some cases, like ToS, a $\delta=0.5$ threshold could be meaningful enough, in other cases, like for the IBM2015 dataset, we preferred to consider in our analysis different values for $\delta$.

\section{Results} \label{sec:results}
In this section, we discuss results based on standard classification metrics, memory usage metrics, and error analysis.

\subsection{Unfairness Detection} \label{sec:results-unfairness-detection}

\begin{table}[!tb]
\centering
\small
    \caption{Classification macro-F1 computed on 10-fold cross-validation for unfair examples on the ToS-30 dataset. We report uniform strategy (U), priority attention-based strategy (Att), and priority loss gain strategy (LG) with sampled memory $\bar{M}$ of size 5. Best results are in bold, second-best results are underlined. \NEW{Standard deviation is reported in subscript format.}}
    \label{tab:tos_performance}
    \scalebox{0.8}{
    \begin{tabular}{cccccc}
    \toprule
    \multicolumn{1}{c}{\textbf{}} & A & CH & CR & LTD & \multicolumn{1}{c}{TER} \\
    \midrule
    \multicolumn{1}{c}{\textbf{No Memory}} & \multicolumn{5}{c}{} \\
    \midrule
    \multicolumn{1}{c}{CNN} & $0.339_{0.089}$ & $0.506_{0.081}$ & $0.403_{0.079}$ & $0.628_{0.065}$ & $0.583_{0.056}$ \\
    \multicolumn{1}{c}{LSTM} & $0.302_{0.073}$ & $0.573_{0.079}$ & $0.363_{0.074}$ & $0.602_{0.056}$ & $0.508_{0.053}$ \\
    \multicolumn{1}{c}{DistilBERT} & $0.447_{0.044}$ & $\mathbf{0.635_{0.047}}$ & $0.620_{0.052}$ & $0.670_{0.042}$ & \underline{$0.748_{0.035}$} \\
    \multicolumn{1}{c}{BERT} & $0.442_{0.050}$ & $0.547_{0.051}$ & $0.653_{0.027}$ & $0.653_{0.029}$ & $0.724_{0.038}$ \\
    \midrule
    \multicolumn{1}{c}{\textbf{Full Memory $M$}} & \multicolumn{5}{c}{\textbf{}} \\
    \midrule
    \multicolumn{1}{c}{MANN (WS)} & $0.483_{0.088}$ & $0.506_{0.083}$ & $0.387_{0.070}$ & $0.635_{0.048}$ & $0.602_{0.062}$ \\
    \multicolumn{1}{c}{MANN (SS)} & $0.465_{0.092}$ & $0.516_{0.078}$ & $0.414_{0.062}$ & $0.605_{0.057}$ & $0.660_{0.051}$ \\
    \multicolumn{1}{c}{MemDistilBERT (WS)} & $0.494_{0.042}$ & $0.565_{0.055}$ & $0.639_{0.039}$ & $0.664_{0.054}$ & $0.705_{0.046}$ \\
    \multicolumn{1}{c}{MemDistilBERT (SS)} & $0.504_{0.045}$ & \underline{$0.609_{0.058}$} & $\mathbf{0.670_{0.041}}$ & $\mathbf{0.686_{0.061}}$ & $0.737_{0.053}$ \\
    \multicolumn{1}{c}{MemBERT (WS)} & $0.477_{0.047}$ & $0.581_{0.043}$ & $0.617_{0.031}$ & $0.650_{0.050}$ & $0.723_{0.055}$ \\
    \multicolumn{1}{c}{MemBERT (SS)} & $\mathbf{0.524_{0.064}}$ & $0.541_{0.056}$ & $0.616_{0.044}$ & $0.630_{0.053}$ & $0.740_{0.048}$ \\
    \midrule
    \multicolumn{1}{c}{\textbf{Sampled Memory $\bar{M}$}} & \multicolumn{5}{c}{} \\
    \midrule
    \multicolumn{1}{c}{MemDistilBERT (WS) (U-5)} & \underline{$0.514_{0.039}$} & $0.556_{0.051}$ & $0.609_{0.042}$ & \underline{$0.678_{0.050}$} & $0.702_{0.044}$ \\
    \multicolumn{1}{c}{MemDistilBERT (WS) (Att-5)} & $0.491_{0.038}$ & $0.559_{0.045}$ & $0.601_{0.047}$ & $0.643_{0.058}$ & $0.703_{0.047}$ \\
    \multicolumn{1}{c}{MemDistilBERT (WS) (LG-5)} & $0.475_{0.037}$ & $0.574_{0.048}$ & \underline{$0.660_{0.045}$} & \underline{$0.678_{0.053}$} & $0.716_{0.048}$ \\
    \multicolumn{1}{c}{MemDistilBERT (SS) (U-5)} & $0.503_{0.041}$ & $0.580_{0.053}$ & $0.617_{0.043}$ & $0.652_{0.055}$ & $0.702_{0.050}$ \\
    \multicolumn{1}{c}{MemDistilBERT (SS) (Att-5)} & $0.448_{0.044}$ & $0.599_{0.051}$ & $0.635_{0.038}$ & $0.661_{0.055}$ & $0.708_{0.057}$ \\
    \multicolumn{1}{c}{MemDistilBERT (SS) (LG-5)} & $0.490_{0.048}$ & $0.536_{0.039}$ & $0.625_{0.044}$ & $0.656_{0.053}$ & $0.706_{0.042}$ \\
    \multicolumn{1}{c}{MemBERT (WS) (U-5)} & $0.514_{0.050}$ & $0.590_{0.056}$ & $0.563_{0.040}$ & $0.677_{0.044}$ & $0.723_{0.045}$ \\
    \multicolumn{1}{c}{MemBERT (WS) (Att-5)} & $0.477_{0.051}$ & $0.597_{0.059}$ & $0.655_{0.056}$ & $0.659_{0.046}$ & $0.710_{0.041}$ \\
    \multicolumn{1}{c}{MemBERT (WS) (LG-5)} & $0.501_{0.048}$ & $0.574_{0.058}$ & $0.625_{0.053}$ & $0.636_{0.047}$ & $\mathbf{0.753_{0.040}}$ \\
    \multicolumn{1}{c}{MemBERT (SS) (U-5)} & $0.463_{0.051}$ & $0.593_{0.056}$ & $0.599_{0.042}$ & \underline{$0.678_{0.045}$} & $0.719_{0.050}$ \\
    \multicolumn{1}{c}{MemBERT (SS) (Att-5)} & $0.459_{0.050}$ & $0.601_{0.050}$ & $0.645_{0.044}$ & $0.649_{0.046}$ & $0.747_{0.051}$ \\
    \multicolumn{1}{c}{MemBERT (SS) (LG-5)} & $0.480_{0.048}$ & $0.580_{0.056}$ & $0.637_{0.040}$ & $0.666_{0.052}$ & $0.723_{0.050}$ \\
    \bottomrule
    \end{tabular}}
\end{table}

\begin{table}[!tb]
\centering
\small
    \caption{Memory usage statistics and metrics concerning predictions on unfair examples only. 
    Activation threshold $\delta$ is set to 0.5.
    Best results are in bold, second-best results are underlined for MRR. Columns C to P@3 are not directly comparable among models due to their different memory usage. \NEW{Standard deviation is reported in subscript format.}}
    \label{tab:tos_30_memory_statistics}
    \scalebox{0.8}{
    \begin{tabular}{ccccccc}
    \textbf{Model} & U & C & CP & P@1 & P@3 & MRR \\
    \toprule
    \multicolumn{6}{c}{Arbitration (A)} \\
    \midrule
    MANN (WS) & $0.311_{0.143}$ & $0.289_{0.114}$ & $0.929_{0.106}$ & $0.571_{0.101}$ & $1.000_{0.083}$ & $0.761_{0.099}$ \\
    MANN (SS) & $0.689_{0.120}$ & $0.644_{0.129}$ & $0.935_{0.093}$ & $0.903_{0.087}$ & $0.968_{0.103}$ & $0.861_{0.089}$ \\
    MemDistilBERT (WS) & $0.489_{0.167}$ & $0.400_{0.149}$ & $0.818_{0.136}$ & $0.273_{0.103}$ & $0.545_{0.088}$ & $0.478_{0.094}$ \\
    MemDistilBERT (SS) & $0.956_{0.035}$ & $0.911_{0.105}$ & $0.953_{0.105}$ & $0.767_{0.087}$ & $0.837_{0.091}$ & $\mathbf{0.864_{0.074}}$ \\
    MemBERT (WS) & $0.756_{0.152}$ & $0.667_{0.121}$ & $0.882_{0.145}$ & $0.147_{0.093}$ & $0.588_{0.105}$ & $0.420_{0.099}$ \\
    MemBERT (SS) & $1.000_{0.000}$ & $0.933_{0.101}$ & $0.933_{0.101}$ & $0.778_{0.084}$ & $0.844_{0.089}$ & \underline{$0.862_{0.061}$} \\
    \midrule
    \multicolumn{6}{c}{Unilateral Change (CH)} \\
    \midrule
    MANN (WS) & $0.169_{0.133}$ & $0.090_{0.135}$ & $0.533_{0.135}$ & $0.000_{0.03}$ & $0.067_{0.02}$ & $0.299_{0.078}$ \\
    MANN (SS) & $0.854_{0.101}$ & $0.730_{0.116}$ & $0.855_{0.109}$ & $0.855_{0.121}$ & $0.961_{0.103}$ & $0.883_{0.091}$ \\
    MemDistilBERT (WS) & $0.404_{0.874}$ & $0.382_{0.109}$ & $0.944_{0.115}$ & $0.250_{0.104}$ & $0.750_{0.090}$ & $0.522_{0.071}$ \\
    MemDistilBERT (SS) & $1.000_{0.001}$ & $0.955_{0.087}$ & $0.955_{0.087}$ & $0.809_{0.107}$ & $0.888_{0.081}$ & \underline{$0.886_{0.069}$} \\
    MemBERT (WS) & $0.801_{0.077}$ & $0.719_{0.105}$ & $0.889_{0.097}$ & $0.222_{0.094}$ & $0.597_{0.084}$ & $0.505_{0.055}$ \\
    MemBERT (SS) & $1.000_{0.008}$ & $0.865_{0.093}$ & $0.865_{0.093}$ & $0.809_{0.080}$ & $0.921_{0.077}$ & $\mathbf{0.889_{0.071}}$ \\
    \midrule
    \multicolumn{6}{c}{Content Removal (CR)} \\
    \midrule
    MANN (WS) & $0.017_{0.011}$ & $0.000_{0.003}$ & $0.000_{0.001}$ & $0.000_{0.001}$ & $0.000_{0.001}$ & $0.335_{0.085}$ \\
    MANN (SS) & $0.672_{0.153}$ & $0.414_{0.149}$ & $0.615_{0.139}$ & $0.282_{0.135}$ & $0.872_{0.098}$ & $0.612_{0.090}$ \\
    MemDistilBERT (WS) & $0.328_{0.127}$ & $0.328_{0.119}$ & $1.000_{0.008}$ & $0.316_{0.113}$ & $0.632_{0.098}$ & $0.478_{0.118}$ \\
    MemDistilBERT (SS) & $1.000_{0.000}$ & $0.948_{0.088}$ & $0.948_{0.088}$ & $0.431_{0.100}$ & $0.879_{0.089}$ & \underline{$0.681_{0.093}$} \\
    MemBERT (WS) & $0.430_{0.135}$ & $0.362_{0.141}$ & $0.84_{0.093}$ & $0.200_{0.114}$ & $0.32_{0.099}$ & $0.328_{0.102}$ \\
    MemBERT (SS) & $1.000_{0.014}$ & $0.914_{0.083}$ & $0.914_{0.083}$ & $0.466_{0.095}$ & $0.879_{0.080}$ & $\mathbf{0.686_{0.096}}$ \\
    \midrule
    \multicolumn{6}{c}{Limitation of Liability (LTD)} \\
    \midrule
    MANN (WS) & $0.037_{0.161}$ & $0.025_{0.155}$ & $0.667_{0.151}$ & $0.330_{0.139}$ & $0.833_{0.124}$ & $\mathbf{0.504_{0.126}}$ \\
    MANN (SS) & $0.814_{0.173}$ & $0.534_{0.169}$ & $0.656_{0.181}$ & $0.313_{0.181}$ & $0.573_{0.141}$ & \underline{$0.501_{0.128}$} \\
    MemDistilBERT (WS) & $0.497_{0.161}$ & $0.416_{0.147}$ & $0.838_{0.133}$ & $0.100_{0.111}$ & $0.275_{0.105}$ & $0.328_{0.104}$ \\
    MemDistilBERT (SS) & $1.000_{0.000}$ & $0.919_{0.114}$ & $0.919_{0.114}$ & $0.224_{0.103}$ & $0.565_{0.091}$ & $0.474_{0.100}$ \\
    MemBERT (WS) & $0.478_{0.138}$ & $0.366_{0.151}$ & $0.766_{0.143}$ & $0.156_{0.139}$ & $0.351_{0.101}$ & $0.303_{0.096}$ \\
    MemBERT (SS) & $1.000_{0.002}$ & $0.919_{0.084}$ & $0.919_{0.084}$ & $0.304_{0.103}$ & $0.609_{0.086}$ & $0.500_{0.089}$ \\
    \midrule
    \multicolumn{6}{c}{Unilateral Termination (TER)} \\
    \midrule
    MANN (WS) & $0.000_{0.000}$ & $0.000_{0.000}$ & $0.000_{0.000}$ & $0.000_{0.000}$ & $0.000_{0.000}$ & $0.499_{0.103}$ \\
    MANN (SS) & $1.000_{0.000}$ & $0.471_{0.120}$ & $0.471_{0.120}$ & $0.438_{0.139}$ & $0.537_{0.114}$ & $0.536_{0.110}$ \\
    MemDistilBERT (WS) & $0.223_{0.130}$ & $0.198_{0.129}$ & $0.889_{0.129}$ & $0.074_{109}$ & $0.074_{0.088}$ & $0.193_{0.090}$ \\
    MemDistilBERT (SS) & $1.000_{0.000}$ & $0.851_{0.091}$ & $0.851_{0.101}$ & $0.438_{0.110}$ & $0.579_{0.108}$ & \underline{$0.567_{0.099}$} \\
    MemBERT (WS) & $0.215_{0.126}$ & $0.198_{0.141}$ & $0.923_{0.137}$ & $0.423_{0.139}$ & $0.538_{0.099}$ & $0.275_{0.089}$ \\
    MemBERT (SS) & $1.000_{0.001}$ & $0.876_{0.094}$ & $0.876_{0.094}$ & $0.380_{0.104}$ & $0.636_{0.101}$ & $\mathbf{0.568_{0.099}}$ \\
    \bottomrule
    \end{tabular}}
\end{table}   

Table~\ref{tab:tos_performance} summarizes the results obtained on the ToS-30 dataset, by reporting the macro-F1 averaged on 10-fold cross-validation.\footnote{For each unfairness category, following~\citep{ruggeri2021detecting} we performed a binary classification task with 10-fold cross-validation: for each category, we report the macro-average over the standard F1 score obtained on the binary task.} A first, clear result is that all BERT models maintain a large performance gap over the other neural baselines. The introduced legal rationales significantly aid memory-augmented models in categories for which there are few positive examples, like A and CR. On the other hand, in category CH we can observe that memory-based approaches do not show an improvement over their counterparts: MANNs perform worse than LSTMs, and MemDistilBERT performs worse than DistilBERT. This may be due to the nature of the natural language rationales given for the CH category, which are probably less informative than those given for the other categories\footnote{Legal rationales for CH category are very similar to each other, with minor distinctions that might be hard to discriminate without supervised guidance.}. The introduction of SS regularization always leads to better performance for MemDistilBERT models with full knowledge. SS regularization is especially beneficial in categories with larger memory sizes like CR, LTD, and TER. Sampling-based models manage to reach performance comparable with their full knowledge counterparts. However, memory sampling attenuates the added contribution of SS regularization, leading to mixed results for different strategies.

Table~\ref{tab:tos_30_memory_statistics} reports memory usage statistics and metrics on unfair clause detection. Results shows that SS brings a crucial contribution in identifying correct explanations: a correct memory slot is ranked in the top-3 positions in over 55-80\% of the cases in all categories. This behavior is reflected also in the MRR score, where SS has a clear advantage over WS.

Many criticisms have recently been raised against the improper use of statistical significance as the only measure to evaluate results in scientific publications~\citep{Amrhein2019ScientistsRU}. However, we also perform the Wilcoxon paired test over the 10-fold cross-validation results, focusing on MemDistilBERT and MANN and the difference between weak and strong supervision. Considering ToS-30, we concur that the results in Table~\ref{tab:tos_performance}, regarding classification performance, are not statistically significant. Yet, regarding performance in terms of correct explanations, reported in Table~\ref{tab:tos_30_memory_statistics}, we observe the following: MemDistilBERT improves over MANN (both with strong supervision) for what concerns coverage precision (CP) on CR, LTD, and TER; coverage (C) on CH, CT, LTD, TER; mean reciprocal ratio (MRR) on TER. Strong supervision outperforms weak supervision (considering DistilBERT) regarding coverage (C) on A, CH, CR, LTD, TER; regarding P@1 on A, CH, TER; regarding P@3 on CR, LTD, TER; regarding MRR on A, CH, LTD, TER. We considered $p$-value $< 0.05$.

\subsection{Claim Detection}

Table~\ref{tab:ibm2015_performance} reports our results on claim detection. Like we did with unfair clause detection, we evaluate knowledge-agnostic neural baselines and MANN models. First of all, we note how MANNs achieved significant improvements over CNNs and LSTMs, suggesting that the introduced knowledge is indeed beneficial to the task itself. Similarly, MemDistilBERT achieves a higher F1-score with respect to standard DistilBERT, whereas MemBERT is comparable with BERT. Interestingly, even the uniform sampling strategy outperforms other neural baselines, suggesting that sampling also compensates for the noise introduced by memories, thus increasing model robustness. This behavior is even more evident with MANNs, where the combination of SS and priority sampling largely outperforms the version with full knowledge. For 1 and 2 topics, MANNs nearly match DistilBERT and BERT, which is remarkable in light of the drastically lower number of parameters. These results support our intuition that controlled, and smart knowledge sampling can be particularly beneficial when limited training data is available.
We observed that with MemDistilBERT, the priority sampling strategies produce quasi-uniform distributions. We think this should be ascribed to the sparsity of relations between memories and examples in this dataset (see Section~\ref{sec:setup:claim-detection}). We thus decided to only run experiments with uniform distribution for BERT.

\begin{table}[!tb]
    \centering
    \small
    \caption{Macro F1-score achieved on IBM2015 dataset. We report uniform strategy (U), priority attention-based strategy (Att), and priority loss gain strategy (LG) with sampled memory $\bar{M}$ of size 10. Best results are in bold, second best results are underlined instead. \NEW{Standard deviation is reported in subscript format.}} 
    \label{tab:ibm2015_performance}
    \scalebox{0.8}{
    \begin{tabular}{ccccc}
        \toprule
         & 1 Topic & 2 Topics & 3 Topics & 4 Topics \\
        \midrule
        \textbf{No Memory} & & & \\
        \midrule
         CNN & $0.196_{0.050}$ & $0.283_{0.055}$ & $0.287_{0.048}$ & $0.268_{0.055}$ \\
         LSTM & $0.194_{0.054}$ & $0.344_{0.059}$ & $0.278_{0.049}$ & $0.272_{0.050}$ \\
         DistilBERT & $0.317_{0.038}$ & $0.431_{0.030}$ & $0.405_{0.042}$ & \underline{$0.451_{0.039}$} \\
         BERT & $0.246_{0.031}$ & $0.442_{0.036}$ & $0.431_{0.038}$ & $0.450_{0.040}$ \\
         \midrule
         \textbf{Full Memory $M$} & & & \\
         \midrule
         MANN (WS) & $0.252_{0.060}$ & $0.380_{0.056}$ & $0.325_{0.055}$ & $0.336_{0.048}$ \\
         MANN (SS) & $0.205_{0.067}$ & $0.392_{0.059}$ & $0.317_{0.601}$ & $0.281_{0.055}$ \\
         \midrule
         \textbf{Sampled Memory $\bar{M}$} & & & \\
         \midrule
         MANN (WS) (U-10) & $0.269_{0.063}$ & $0.406_{0.051}$ & $0.331_{0.058}$ & $0.355_{0.056}$ \\
         MANN (WS) (Att-10) & $0.251_{0.061}$ & $0.402_{0.050}$ & $0.322_{0.055}$ & $0.358_{0.058}$ \\
         MANN (WS) (LG-10) & $0.259_{0.059}$ & $0.408_{0.053}$ & $0.332_{0.055}$ & $0.340_{0.051}$ \\
         MANN (SS) (U-10) & $0.297_{0.060}$ & $0.400_{0.055}$ & $0.319_{0.059}$ & $0.352_{0.056}$ \\
         MANN (SS) (Att-10) & $0.264_{0.064}$ & $0.423_{0.059}$ & $0.332_{0.062}$ & $0.348_{0.057}$ \\
         MANN (SS) (LG-10) & $0.302_{0.068}$ & $0.424_{0.060}$ & $0.344_{0.061}$ & $0.354_{0.059}$ \\
         MemDistilBERT (WS) (U-10) & $0.311_{0.035}$ & \underline{$0.457_{0.036}$} & $\mathbf{0.454_{0.041}}$ & \underline{$0.453_{0.043}$} \\
         MemDistilBERT (WS) (Att-10) & $0.275_{0.042}$ & $0.449_{0.041}$ & $0.422_{0.039}$ & $0.434_{0.039}$ \\
         MemDistilBERT (WS) (LG-10) & $0.305_{0.045}$ & $0.449_{0.038}$ & $0.428_{0.036}$ & $0.428_{0.040}$ \\
         MemDistilBERT (SS) (U-10) & \underline{$0.341_{0.044}$} & $0.442_{0.043}$ & \underline{$0.444_{0.037}$} & $0.436_{0.039}$ \\
         MemDistilBERT (SS) (Att-10) & $\mathbf{0.354_{0.045}}$ & $0.424_{0.040}$ & $0.421_{0.038}$ & $0.423_{0.047}$ \\
         MemDistilBERT (SS) (LG-10) & $0.290_{0.050}$ & $\mathbf{0.459_{0.051}}$ & $0.411_{0.049}$ & $0.444_{0.049}$ \\
         MemBERT (WS) (U-10) & $0.278_{0.039}$ & $0.430_{0.040}$ & $0.427_{0.039}$ & $0.428_{0.041}$ \\
         MemBERT (WS) (Att-10) & $0.287_{0.040}$ & $0.424_{0.039}$ & $0.441_{0.043}$ & $0.432_{0.044}$ \\
         MemBERT (WS) (LG-10) & $0.261_{0.042}$ & $0.419_{0.041}$ & $0.432_{0.042}$ & $\mathbf{0.456_{0.043}}$ \\
         MemBERT (SS) (U-10) & $0.253_{0.043}$ & $0.421_{0.039}$ & $0.436_{0.040}$ & $0.432_{0.038}$ \\
         MemBERT (SS) (Att-10) & $0.260_{0.045}$ & $0.402_{0.044}$ & $0.425_{0.049}$ & $0.442_{0.040}$ \\
         MemBERT (SS) (LG-10) & $0.268_{0.047}$ & $0.422_{0.046}$ & $0.425_{0.051}$ & $0.431_{0.048}$ \\
         \bottomrule
    \end{tabular}}
\end{table}

\begin{figure}[!t]
    \centering
    \resizebox{\columnwidth}{!}{%
    \includegraphics[clip, trim=0cm 8cm 4.5cm 2.5cm, page=1]{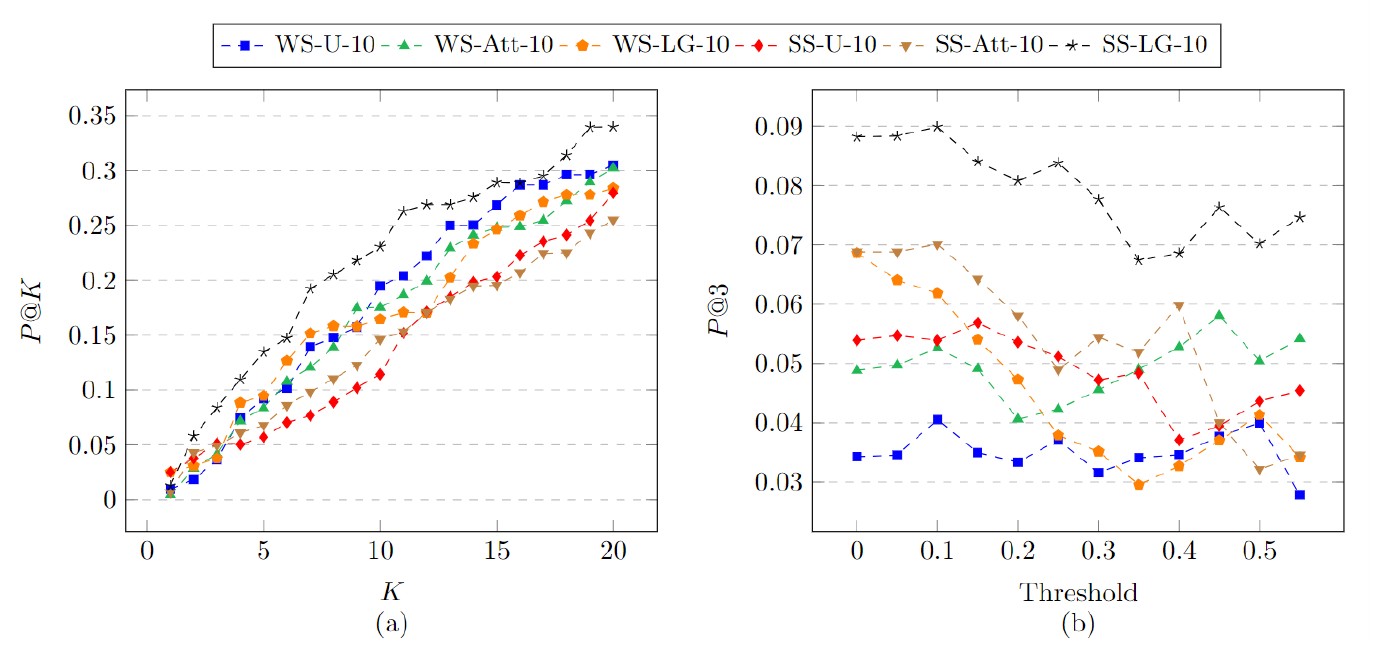}%
    }
    \caption{MemDistilBERT analysis on IBM2015, 1-Topic. (a) P@K for increasing $K$ values and $\delta = 0.25$; (b) P@3 for increasing $\delta$ values. Metrics for sampling-based models are averaged across three distinct inferences on test set.}
    \label{fig:memory_graphs}
\end{figure}

Differently from unfairness detection, the large memory size hinders selective memory lookup operations. Additionally, claim detection on the IBM2015 dataset is a challenging task where existing solutions reach comparably low performance~\citep{levy-etal-2014-context,DBLP:conf/ijcai/LippiT15}. For these reasons, following previous work on the same dataset, we focus our analysis on retrieval metrics and compute them over incremental values of activation threshold $\delta$. In particular, we mainly focus on the P@K metric, since it is a valuable indicator of the model retrieval capabilities.\footnote{For example, \citep{levy-etal-2014-context} reports P@200 on a slightly different dataset version.} Figure~\ref{fig:memory_graphs}(a) reports values of P@K on the 1-Topic case, for incremental values of $K$, with fixed $\delta = 0.25$, whereas Figure~\ref{fig:memory_graphs}(b) reports P@3 with incremental $\delta$ values. 
In both cases, we observe a slight advantage in using loss gain priority sampling (LG-10) coupled with SS.

\subsection{Error Analysis and Discussion}

As observed, in unfairness detection, SS regularization is highly effective in selecting the appropriate legal rationales when detecting an unfair clause (see Table~{\ref{tab:tos_30_memory_statistics}}). This is especially true with categories with a larger memory like CR, LTD, and TER. Compared to WS, SS regularization helps in filtering out irrelevant legal rationales. For instance, consider the following clause from Linden Lab contract, which is unfair according to the limitation of liability (LTD) category:
\begin{quote}
    \emph{In such event, you will not be entitled to compensation for such suspension or termination, and you acknowledge that Linden Lab will have no liability to you in connection with such suspension or termination.}
\end{quote}
\noindent BERT fails in identifying the clause as being unfair, whereas MemBERT (WS) and MemBERT (SS) correctly detect the unfairness. Nonetheless, MemBERT (WS) almost uses half of the memory slots (17 out of 28). In contrast, MemBERT (SS) selects only four of them and shows a preference for the target memory slot, which corresponds to the following rationale:

\begin{quote}
    \emph{Since the clause states that the provider is not liable for any technical problems, failure, suspension, disruption, modification, discontinuance, unavailability of service, any unilateral change, unilateral termination,  unilateral limitation including limits on certain features and services, or restriction to access to parts or all of the Service without notice.}
\end{quote}
Section~II in the Supplementary Material contains other interesting illustrations: a model with WS failing to use the memory while SS regularization overcomes this issue, and a model with SS showing a correct preference for a target memory while still using a large number of memory slots.

We notice that the impact of SS regularization is more significant when the memory is limited in size and each memory slot is associated with several input samples. This is not the case with claim detection, where memory slots are in the hundreds and memory/claim associations are few. For instance, consider the following sentence containing a claim:
\begin{quote}
    \emph{The argument from parsimony (using Occam's razor) contends that since natural (non-supernatural) theories adequately explain the development of religion and belief in gods, the actual existence of such supernatural agents is superfluous and may be dismissed unless otherwise proven to be required to explain the phenomenon.}
\end{quote}

The claim is associated with the following target evidence:
\begin{quote}
    \emph{Sigmund Freud stressed fear and pain, the need for a powerful parental figure, the obsessional nature of ritual, and the hypnotic state a community can induce as contributing factors to the psychology of religion.}
\end{quote}

BERT fails to classify the sentence as containing a claim, whereas MemBERT (WS) and MemBERT (SS) correctly perform the classification. Nonetheless, neither of the MemBERT models are able to select the target evidence. We observe that MemBERT models capture the semantic of the claim but focus on religion-centered evidence from articles other than the one containing the above target evidence. For instance, MemBERT (SS) selects as the best candidate the following evidence:
\begin{quote}
    \emph{The Catholic Church, following the teachings of Saint Paul the Apostle, Saint Thomas Aquinas, and the dogmatic definition of the First Vatican Council, affirms that God's existence can be known with certainty from the created world by the natural light of human reason}
\end{quote}

However, MemBERT (SS) fails to capture the stance of the claim sentence with respect to the selected evidence. We believe that the complexity and length of considered texts hinder the effectiveness of a successful evidence retrieval based on semantic similarity. The reported example also suggests that multiple evidence from different articles may be equivalently acceptable as support to a given claim. This fact inherently increases the complexity of the classification task on the IBM2015 dataset. Section~III in the Supplementary Material contains other interesting illustrations: models with WS and SS selecting memory slots to correctly classify a sentence containing a claim despite the lack of target evidence, and a model with SS that uses fewer memory slots than its WS counterpart.
Like with unfairness detection, categories A and CR, we observe a performance improvement when the amount of training data is limited.

The introduction of smart sampling strategies allows scaling to larger memories. However, the introduced priority-based strategies lack proper conditioning on the input, but rather learn the dataset-level importance of each memory slot. This works well in unfairness detection, where there are few memory slots and they are associated with many samples; not so well in claim detection, where there are many memory slots, each associated with one or very few samples. In our experiments, this led priority-based strategies to learn a quasi-uniform priority distribution. This suggests a possible direction for improving the adoption of input-conditioned sampling strategies compatible with SS regularization. For instance, a parametric model could be trained to embed input texts near their target memory slots into a latent space~{\citep{munkhdalai2019metalearned}}. 

\section{Conclusions} \label{sec:conclusions}

In many application domains, explainability is a crucial requirement, explanations can be constructed from expert knowledge, and expert knowledge is easier to acquire in plain natural language than in a structured format.
For this reason, we studied an extension of 
transformers 
whereby
memories are used to store possible natural language explanations for the model outputs. 
We validated our proposal in two challenging domains: (i) providing justifications for unfair clause detection in the legal domain and (ii) identifying evidence as support for claim detection in the context of argument mining.
We obtained interesting results not only in terms of interpretability, but also, remarkably, in terms of classification metrics, showing that it is possible to make models interpretable without necessarily losing in performance. Most notably, when strong supervision is applied, memory-augmented models learn to give more importance to target memory slots and, in most cases, yield higher classification performance. Memory sampling allows scaling up to large knowledge bases without an appreciable performance loss. However, sampling strategies are less effective with high number of memories and comparatively less associations in the training set.
\NEW{In contrast, when considering a relatively small knowledge base, as in the legal scenario, both priority sampling strategies are superior to random. 
However, their effectiveness is sensitive to the textual representation of the knowledge base, and no strategy is a clear winner over the other across all the evaluated scenarios.}

Possible improvements could be obtained by introducing sampling strategies explicitly conditioned on the given input, or priority-based sampling strategies that integrate memory-related regularizations such as strong supervision. By doing so, the sampling strategies could learn to take into account different factors like the similarity between the input and the memory slots, the impact of each memory slot on task performance, and the existence of target memory slots for a given input. By balancing these factors during learning, one could modify the content of each memory slot dynamically instead of relying on static memory content, like we do. Adaptive data-driven knowledge integration for example has been explored by~\citep{hu-etal-2016-deep}.
Further performance improvement may come from more sophisticated memory look-up and reasoning mechanisms. We plan to investigate all these in future work.
Another interesting direction would be 
the application of our method
to natural language generation tasks. In this way, one could generate text by leveraging the knowledge contained in the memories. For example, arguments could be generated based on available evidence.

\section*{Acknowledgment}
This work was partially supported by TAILOR, project funded by EU Horizon 2020 research and innovation programme under GA No 952215.

\section*{CRediT}

\textbf{Federico Ruggeri:} Conceptualization; Formal analysis; Investigation; Methodology; Software; Writing- original draft; Writing- review \& editing; Validation;  \textbf{Marco Lippi:} Formal analysis; Funding acquisition; Writing- review \& editing; Supervision \textbf{Paolo Torroni:} Formal analysis; Funding acquisition; Writing- review \& editing; Supervision.

\appendix

\bibliographystyle{elsarticle-num}
\bibliography{bibliography}

\end{document}